# Collision-Free Kinematics for Redundant Manipulators in Dynamic Scenes using Optimal Reciprocal Velocity Obstacles*

Liangliang Zhao[1], Jingdong Zhao[1], Hong Liu[1], and Dinesh Manocha[2]

*Abstract—* We present a novel algorithm for collision-free kinematics of multiple manipulators in a shared workspace with moving obstacles. Our optimization-based approach simultaneously handles collision-free constraints based on reciprocal velocity obstacles and inverse kinematics constraints for high-DOF manipulators. We present an efficient method based on particle swarm optimization that can generate collision-free configurations for each redundant manipulator. Furthermore, our approach can be used to compute safe and oscillation-free trajectories in a few milli-seconds. We highlight the real-time performance of our algorithm on multiple Baxter robots with 14-DOF manipulators operating in a workspace with dynamic obstacles. Videos are available at https://sites.google.com/view/collision-free-kinematics

## I. INTRODUCTION

Redundant manipulators are widely used in complex and cluttered environments for various kinds of tasks. Moreover, many applications use multiple manipulators that work cooperatively in an environment with moving obstacles (e.g. humans). To perform the tasks, the planning algorithms must compute trajectories that simultaneously satisfy two sets of constraints:

1. Inverse Kinematics Constraints: Given the pose of each end-effector, we need to compute joint coordinates that satisfy kinematics constraints [1].
2. Collision-free Constraints: The manipulators should not collide with each other or with the static or dynamic obstacles in the environment [2].

There is considerable prior work on computing efficient inverse kinematics solutions using numerical or optimization methods for redundant manipulators and good software packages are widely available [4-9]. While these methods may account for self-collisions between different links of a manipulator, they do not consider collisions with other manipulators or dynamic obstacles. Similarly, there is a large body of work on collision-avoidance for multi-robot systems or dynamic obstacles. These works are mostly limited to rigid bodies or only take dynamics constraints into account (e.g., maximum acceleration or non-holonomic) [21-31]; they have not been used for high-DOF manipulators.

**Main Results**: We present a real-time algorithm to compute a collision-free inverse kinematics solution for each redundant manipulator that operates in a three-dimensional

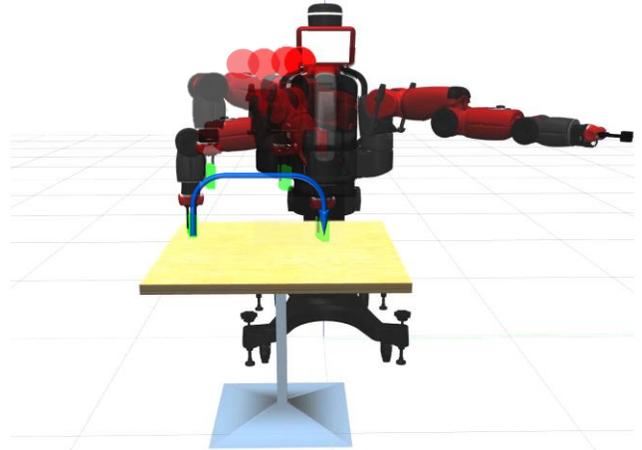

Fig 1. Solving inverse kinematics on a Baxter robots manipulator (7-DOF) in the workspace. Our algorithm can compute the inverse kinematics solutions for these redundant manipulators, and avoid collisions (red sphere) between the robots. The blue line is the desired path of the end effector.

workspace among other redundant manipulators and static or dynamic obstacles.

Our approach is based on an efficient optimization-based technique that uses a novel combination of Inverse Jacobian methods, sequential quadratic programming (SQP), and reciprocal velocity obstacles (RVO). We extend the formulation of RVO [24] to redundant manipulators and pose the trajectory computation problem as a high dimensional optimization problem with constraints. We present an efficient method to solve this problem using particle swarm optimization (PSO). The basic idea is that we compute inverse kinematics solutions that satisfy the fitness function value of PSO, while accounting for collision-avoidance constraints. Our algorithm initially uses a parallel inverse kinematic method (TRCK-IK) for each redundant manipulator in the workspace and builds a swarm of candidate particles for PSO. After PSO initialization, we continue to search for optimized inverse kinematics solutions by updating each particle. The RVO constraints of a redundant manipulator are used to check for collisions in the given time window. These collision avoidance constraints are formulated using linear inequalities and taken into account to compute the fitness values of the PSO's particles. The movements of each particle are guided by the fitness values to be optimized. In order to guarantee that

1. State Key Laboratory of Robotics and System, Harbin Institute of Technology, Harbin, 150001, China. (e-mail: zhaojingdong@hit.edu.cn).
2. Department of Computer Science and Electrical & Computer Engineering, University of Maryland at College Park, MD, USA.

there is no collision, we use Gazebo simulator that supports physics engines (include Open Dynamics Engine, Bullet, DART and Simbody) and offers the ability to efficiently and accurately simulate robots in dynamic environments.

We have evaluated our method in environments with multiple Baxter robots and computed collision-free inverse kinematics solutions that satisfy the kinematic and RVO constraints of multiple Baxter arms (each arm has 7-DOF). We also account for dynamic obstacles moving in the working environment. In general, the run time to generate a collision-free inverse kinematics solution for each redundant manipulator in a common workspace depends on the kinematic solver and the number of iterations of the PSO method. In our benchmarks, it takes a few milliseconds on a single CPU core. Furthermore, the complexity increases almost linearly with the number of manipulators or dynamic obstacles.

The rest of the paper is organized as follows. In Section II, we briefly discuss related work in inverse kinematics and the RVO algorithm. In Section III, we give an overview of the inverse kinematics algorithm TRAC-IK, including the inverse Jacobian algorithm and the SQP algorithm. In Section IV, we extend the notion of velocity obstacles to redundant manipulators. We present a novel optimization-based PSO algorithm in Section V that accounts for inverse kinematics and RVO constraints. In Section VI, we highlight the performance on the arms of the Baxter robot in dynamic environments with multiple manipulators and obstacles.

## II. Previous Work

In this section, we give a brief overview of prior work on collision-free navigation of multiple robots and inverse kinematics of redundant manipulators.

Some of the widely used methods to generate valid inverse kinematics solutions are based on Inverse Jacobian methods. The Kinematics and Dynamics Library (KDL), distributed by the Orocos Project, can compute forward position kinematics to inverse kinematics based on the Inverse Jacobian method [4]. Beeson and Ames [5] propose an inverse kinematics algorithm called TRAC-IK that can improve several failure points of KDL. Starke et al. [6] present a biologically-inspired method for solving the inverse kinematics problem of fully-constrained robot geometries. Based on cyclic coordinate descent (CCD) and natural-CCD, Andrés et al. [7] present an algorithm to solve the inverse kinematics problem of hyper-redundant and soft manipulators. Stilman et al. [8] introduce a unified representation for task space constraints by global randomized joint space path planning. Marcos et al. [9] introduce a method that combines the closed-loop pseudo-inverse method with a multi-objective genetic algorithm to solve the inverse kinematics of redundant manipulators. Other techniques are based on algebraic solvers [10]. None of these methods consider collision-free constraints or dynamic obstacles.

Various shape trajectory control approaches can achieve good performances in terms of computing an inverse kinematics solution of the redundant manipulator. These approaches include a spatial curve based on a tractrix curve [11], shape trajectory data from sidewinder rattlesnakes [12-14], mechanics modeling [15], plain spline fitting and extended spline fitting methods [16], the curvature gradient of a constant parameter along the segment arm [17], physical curves [18], a backbone curve [19], and an inchworm step [20].

There is considerable work on collision-free navigation of multiple robots sharing a common workspace. Some of the widely used solutions are based on velocity obstacles [21-23]. Berg et al. [24] propose the Reciprocal Velocity Obstacle algorithm for real time multi-agent navigation. This algorithm can be used to generate smooth paths for agents moving in the same environment and has been extended to handle bounds on acceleration [25] or simple airplanes in 3D [26].

Other velocity-obstacle-based methods account for dynamic constraints. These include differential-drive [27], double integrator [28], arbitrary integrator [29], car-like robots [30], linear quadratic regulator (LQR) controllers [31], non-linear equations of motion [32], etc. Some other algorithms like NH-ORCA [28] transfer non-linear equations of motion into a linear formulation. However, these methods are not designed for high-DOF redundant manipulators.

## III. Inverse Kinematics Algorithm

In this section, we describe the characteristics of the inverse kinematics solver that is used in our algorithm. The functional form of the inverse kinematics problem is given by:

$$\theta_{1,...,n} = f^{-1}(\xi_E) \qquad (1)$$

where $\xi_E$ is the desired pose of the end effector and $\theta_{1,...,n}$ are required joint coordinates. Because of redundancy, it is necessary to consider a numerical method that compute a feasible solution. The inverse kinematics solver TRAC-IK is a parallel method [6] that combines two inverse kinematics methods, including a Newton-based convergence algorithm (KDL) and an SQP approach. It performs a parallel search using these methods and terminates when either of these algorithms converges to an inverse kinematics solution. The most common values of a seed joint $\theta_{seed}$ for the Inverse Jacobian algorithm and the SQP algorithm are the current joint values. When all the elements fall below a stopping criterion, the current joint vector $\theta$ is returned as an inverse kinematics solution.

*a) KDL*: Using a singular value decomposition, the Moore-Penrose pseudoinverse of the Jacobian $J^\dagger$ is computed to translate the partial derivatives in the joint space to the Cartesian space. Next, an inverse kinematics solution is computed by iterating the function

$$\begin{bmatrix} \theta_{1(k+1)} \\ \vdots \\ \theta_{n(k+1)} \end{bmatrix} = \begin{bmatrix} \theta_{1(k)} \\ \vdots \\ \theta_{n(k)} \end{bmatrix} + J^\dagger \begin{bmatrix} Err_1 \\ \vdots \\ Err_n \end{bmatrix} \qquad (2)$$

where $\theta_{i(k+1)} (i = 1, ..., n)$ are the current joint values of the manipulator and $Err_i$ is the Cartesian error of the end effector, which can be computed by the previous joint values $\theta_{i(k)}$.

*b) SQP*: This algorithm (as introduced in [5]) considers an inverse kinematics problem with the following form:

$$\arg\min_{\theta \in \mathbb{R}^n} (\theta_{seed} - \theta)^T (\theta_{seed} - \theta) \qquad (3)$$
$$s.t. \quad g(\theta) \leq b,$$

where $\theta_{seed}$ is the $n$-dimensional seed value of the joints and $g(\theta)$ corresponds to the constraints, including the value limits of each joint, the Euclidean distance that needs to be satisfied for the required end-effector position, and the angular distance error.

## IV. RECIPROCAL VELOCITY OBSTACLES FOR HIGH-DOF MANIPULATORS

In this section, we present our method for collision avoidance, which extends the notion of RVO to redundant manipulators based on velocity obstacles constraints. We derive the formulation for calculating the RVO constraints while they are induced by other redundant manipulators and dynamic obstacles in a time window. We also present an efficient technique to compute the velocity of each link based on the inverse kinematics solutions of the redundant manipulator.

In order to apply RVO during collision avoidance, each movable link of the redundant manipulator is represented into a series of spheres, and each movable joint is described by a sphere, as shown in Fig. 2. The number of spheres and their relative positions and radii are dynamically determined by the size of the links and joints. Furthermore, we make sure that there are no collisions by choosing bounding sphere. Also, we can choose different number of spheres based on the environments and task requirements. In this paper, we assume that we know the number of spheres and their relative positions and radii that used for each link of the manipulator (from $S_1$ to $S_7$ in Fig. 2). These spheres move around other manipulators and dynamic obstacles. Each dynamic obstacle is also assumed to be a sphere or is bounded by a sphere (Obstacle 1 and 2 in Fig. 2). In practice, our collision avoidance method tends to be conservative because of these bounding sphere approximations.

### A. Reciprocal Velocity Obstacles

In the Cartesian space, let $S_m$ ($m = 1, …, M$, where $M$ is the number of spheres that are used for this manipulator's decomposition) be one of the spheres on the link of the redundant manipulator and let $A$ be an obstacle. As shown in Fig. 3, the sphere $S_m$ is centered at $O_m = (o_{x_m}, o_{y_m}, o_{z_m})$ with radius $r_m$, and the obstacle $A$ are centered at $O_A = (o_{x_A}, o_{y_A}, o_{z_A})$ with radii $r_A$. The length of the vector between the two centers can be defined as:

$$D_m = \sqrt{(o_{x_A} - o_{x_m})^2 + (o_{y_A} - o_{y_m})^2 + (o_{z_A} - o_{z_m})^2} \quad (4)$$

If $D_m \leq r_m + r_A$, we conclude that sphere $S_m$ and $A$ are colliding.

The Minkowski sum of these spheres $S_m$ and $A$ can be described by the equation

$$S_m \oplus A = \{s_m + a \mid s_m \in CH(S_m), a \in CH(A)\} \quad (5)$$

where $CH(S_m)$ and $CH(A)$ are the convex hulls of two spheres $S_m$ and $A$, respectively. The RVO for sphere $S_m$ induced by sphere $A$ for time window $\tau$ is given as:

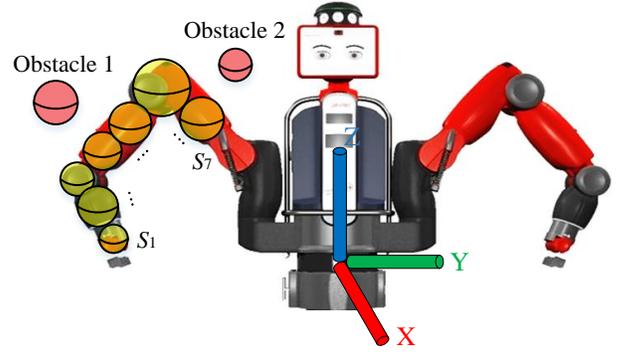

Fig 2. The right arm of Baxter robot is decomposed into a series of spheres. The combined velocity obstacle for the sphere (red) is the union of the individual velocity obstacles of the other spheres. This way, we can provide collision-avoidance guarantees.

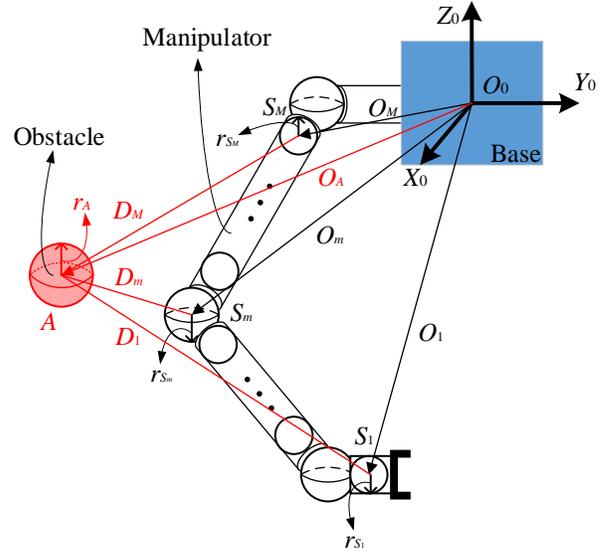

Fig 3. A manipulator and a dynamic obstacle are bounded by spheres whose centers are $O_1, …, O_m$ and $O_A$ and radii are $r_{S_1},…, r_{S_M}$ and $r_A$, respectively.

$$RVO^\tau_{S_m|A} = \{v \mid \lambda^\tau(O_m, v - v_A) \cap A \oplus -S_m \neq \varnothing\} \quad (6)$$

where $-S_m = \{-s_m \mid s_m \in CH(S_m)\}$, $v_A$ is the velocity vector of the sphere $A$, and $\lambda^\tau(O_m, v - v_A)$ is a ray starting at $(o_{x_m}, o_{y_m}, o_{z_m})$ with direction $v$,

$$\lambda^\tau(O_m, v - v_A) = \{O_m + t(v - v_A) \mid t \in [0, \tau]\} \quad (7)$$

As shown in Fig. 4, if the sphere $S_m$ has velocity $v_m$, we observe that sphere $S_m$ may collide with sphere $A$ during the time interval $[0, \tau]$ if the relative velocity vector of $v_m - v_A$ is inside the region $RVO^\tau_{S_m|A}$. To avoid a possible collision before time $\tau$, the relative velocity vector of $v_m - v_A$ must be outside $RVO^\tau_{S_m|A}$.

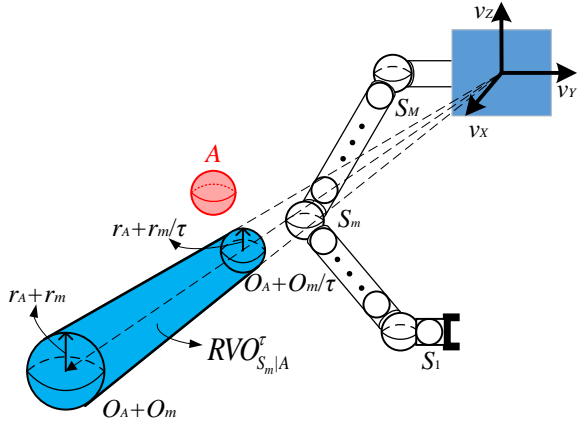

Fig 4. The shaded area (blue) represents the velocity obstacles for the sphere $A$ induced by the sphere $S_m$ in time window $\tau$ in the three-dimensional workspace.

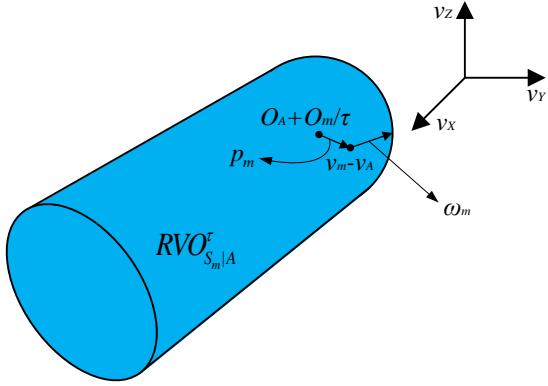

Fig 5. In the xyz-dimensional, the vector $p_m$ is the factor which determine if there is a collision, and the vector $\omega_m$ is the factor which informally represents the amount of load the sphere takes for collision avoidance.

### B. RVO Constraints

As Section IV-A described, the formulation of RVO constraints can be used to navigate a sphere in the dynamic workspace without collisions. To select a velocity for sphere $S_m$, we introduce a constraint defined with respect to the velocity $v_m - v_A$ and the region $RVO^\tau_{S_m|A}$.

Having identified a velocity $v_m$ for sphere $S_m$, the vector $\omega_m$ from $v_m - v_A$ to the closest point of the boundary of the $RVO^\tau_{S_m|A}$, as shown in Fig. 5, is defined as follows:

$$\omega_m = (\arg\min \| v - (v_m - v_A) \|_2) - (v_m - v_A), v \in \partial RVO^\tau_{S_m|A} \tag{8}$$

where $\partial RVO^\tau_{S_m|A}$ is the boundary of the velocity obstacle. Let $p_m$ represent the vector from point $O_m - O_A/\tau$ to $v_m - v_A$. As with previous formulations, the constraint factor $\psi_m$ is defined as follows:

$$\psi_m = \frac{p_m}{|p_m|} \cdot \omega_m \tag{9}$$

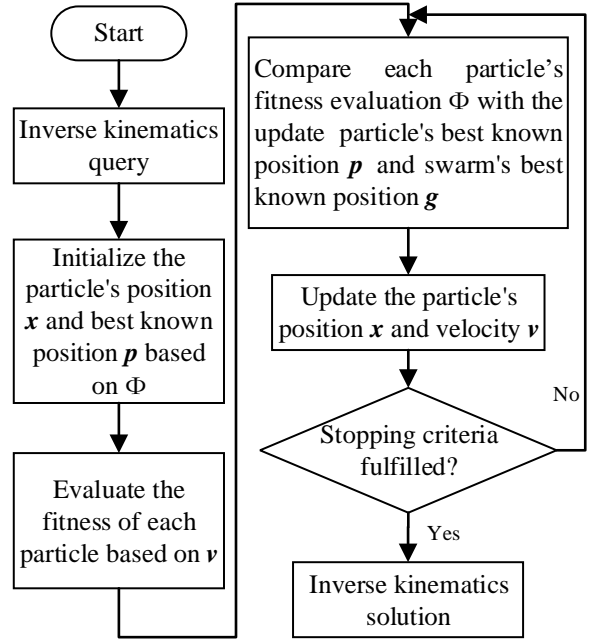

Fig 6. A schematic overview of our approach for finding the inverse kinematics solutions of redundant manipulators based on PSO.

If $\psi_m \leq 0$, then the sphere $S_m$ will collide with the obstacle sphere $A$. If $\psi_m > 0$, there will be no such collision within the time interval $[0, \tau]$. In the next section, we use this constraint factor to combine the inverse kinematics and collision avoidance constraints.

## V. Relating Inverse Kinematics Algorithm to RVO Based on Particle Swarm Optimization

As Section III described, we can compute the inverse kinematics solutions using an inverse kinematics algorithm. Based on the inverse kinematics solutions, we can compute the constraint factor $\psi_m$ (as Equations (9) described) of the spheres in the redundant manipulator and use these values to find a better seed joint $\theta_{seed}$ for the inverse kinematics algorithm to perform obstacle avoidance. Secondary constraints of angular distance error $E$ of the manipulator are used to account for slight changes to the joints of the redundant manipulator.

We use a PSO algorithm to optimize continuous nonlinear functions. As Sections III and IV describe, inverse kinematics and reciprocal velocity obstacles constraints represent a continuous optimization problem. Therefore, we can use PSO to find an inverse kinematics solution without any collisions with the obstacles. The overall steps of the inverse kinematics method are shown in Fig. 6, which include inverse kinematics solver and the reciprocal velocity obstacles constraints combined with PSO.

### A. Particles Initialization

A basic variant of the PSO algorithm is a swarm of candidate particles. These particles tend to move around in the search space according to the fitness function. The movements of the particles are guided by the best known positions of individual particles in the search space and by the entire swarm's best-known position.

In this paper, we assume that the number of DOF of all manipulators in the workspace is fixed and denoted as $n$. The PSO consists of particles $x_{ms}$ for $ms = 1, \ldots, N$ represents the number of particles in the swarm. Thus, we can encode the joint variable configuration $\theta_{seed}$ as the particle $x_{ms}$ for the individuals,

$$\theta_{seed} \triangleq x_{ms} = (x_{ms,1}, \ldots, x_{ms,n}) \quad (10)$$

where $\theta_{seed}$ is the seed value of the joints for inverse kinematics algorithm (described in Section III). Each of the particle has a position $(x_{ms,1}, \ldots, x_{ms,n}) \in \mathbb{R}^n$ in the search space (caused by the joint limits of each manipulator) and the velocity $v_{ms} = (x_{ms,1}, \ldots, x_{ms,n}) \in \mathbb{R}^n$. During the iterative computations of the IK algorithm, the joint limit constraints are used to ensure that the position remains within the fundamental limits of the manipulator. These limits can be defined as:

$$\theta_{i\min} \leq x_{ms,i} \leq \theta_{i\max}, i = 1, \ldots, n \quad (11)$$

### B. Fitness Function

To avoid collisions with the multiple moving obstacles in dynamic environments, the constraint factor $\psi_{ms}$ of RVO is introduced in the fitness function $\Phi_{ms}$ of the particles $x_{ms}$, which must be minimized to measure the fitness of a particle. In addition to finding a smoother motion, the angular distance error $E_{ms}$ generates an inverse kinematics solution that bears a relation to the current joint values. We use a method that aims to select a solution that minimizes the value of the fitness function $\Phi_{ms}$.

$$\Phi_{ms} = \alpha \psi_{ms} + \beta E_{ms} \quad (12)$$

$$\psi_{ms} = \sum_{m=1}^{M} \psi_m = \sum_{m=1}^{M} \frac{p_m}{|p_m|} \cdot \omega_m \quad (13)$$

$$E_{ms} = \sum_{i=1}^{n} \lambda_i \left| (\theta_i - \theta_i^c) \right| \quad (14)$$

where $\psi_{ms}$ is the constraint factor computed by summing the RVO constraint of all spheres in the workspace, $E_{ms}$ is the angular distance error, $\alpha$ and $\beta$ are the constants used to regulate the variation range of $\psi_{ms}$ and $E_{ms}$, $\lambda_i$ is the weight of each joint angle, $\theta_i$ is the inverse kinematics solution of all manipulators, and $\theta_i^c$ is the joint variable of the current configuration.

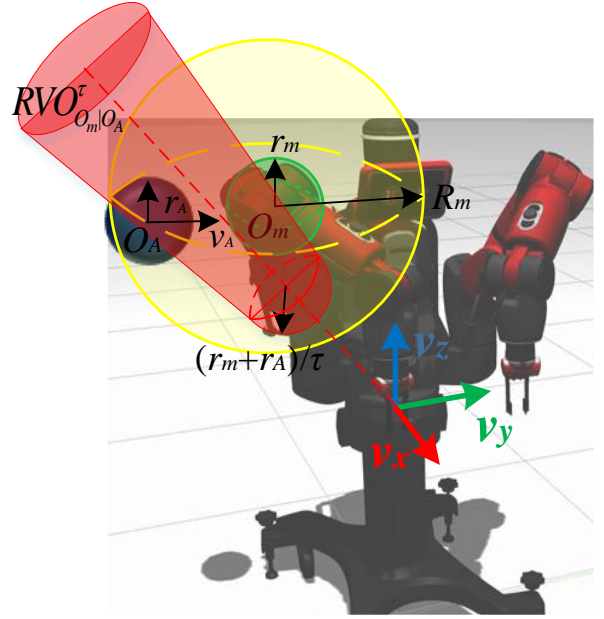

Fig 7. The general method of the combined TRAC-IK and RVO for the right arm of Baxter. The three-dimensional working environment contains a dynamic obstacle (blue sphere), which is moving in the environment. The red area represents the velocity obstacle $RVO^{\tau}_{O_m|O_A}$ for the sphere $O_m$ (green sphere) induced by the sphere $A$ (blue sphere) in the time window $\tau$ in the three-dimensional workspace. The yellow sphere represents the neighbor region $R_m$.

### C. Particle Swarm Optimization

After PSO is initialized with a group of particles and a fitness function, it continues to search for optimal solution by updating particles $x_{ms}$. Based on the fitness function value $\Phi_{ms}$, let $p_{ms}$ be the best known position of particle $x_{ms}$, and let $g$ be the best known position obtained so far of any particle in the swarm. During each iteration, each particle $x_{ms}$ is updated by $p_{ms}$ and $g$. Each particle updates its velocity and positions with Equations (15) and (16).

TABLE I. INVERSE KINEMATICS RESULTS OF REDUNDANT MANIPULATORS USING A SINGLE CPU CORE 2.81 GHZ INTEL I7-7700HQ

| DOF ($n$) | Environment | Number of spheres | Time window $\tau$ (s) | PSO Iterations ($T$) | Number of particles ($N$) | Processing time for a solution (all arms) (ms) |
|---|---|---|---|---|---|---|
| 7 | One Baxter Robot arm and one dynamic obstacle | 8 | 5 | 2 | 2 | 2.01 |
| 14 | Both Baxter Robot arms | 14 | 5 | 3 | 2 | 5.66 |
| 14 | Both Baxter Robot arms and one dynamic obstacle | 15 | 5 | 3 | 3 | 9.12 |
| 14 | Both Baxter Robot arms and two dynamic obstacle | 16 | 5 | 4 | 3 | 12.4 |
| 14 | Both Baxter Robot arms and three dynamic obstacle | 17 | 5 | 4 | 4 | 19.2 |
| 28 | Two Baxter Robots | 28 | 5 | 3 | 3 | 20.4 |
| 42 | Three Baxter Robots | 42 | 5 | 4 | 3 | 45 |

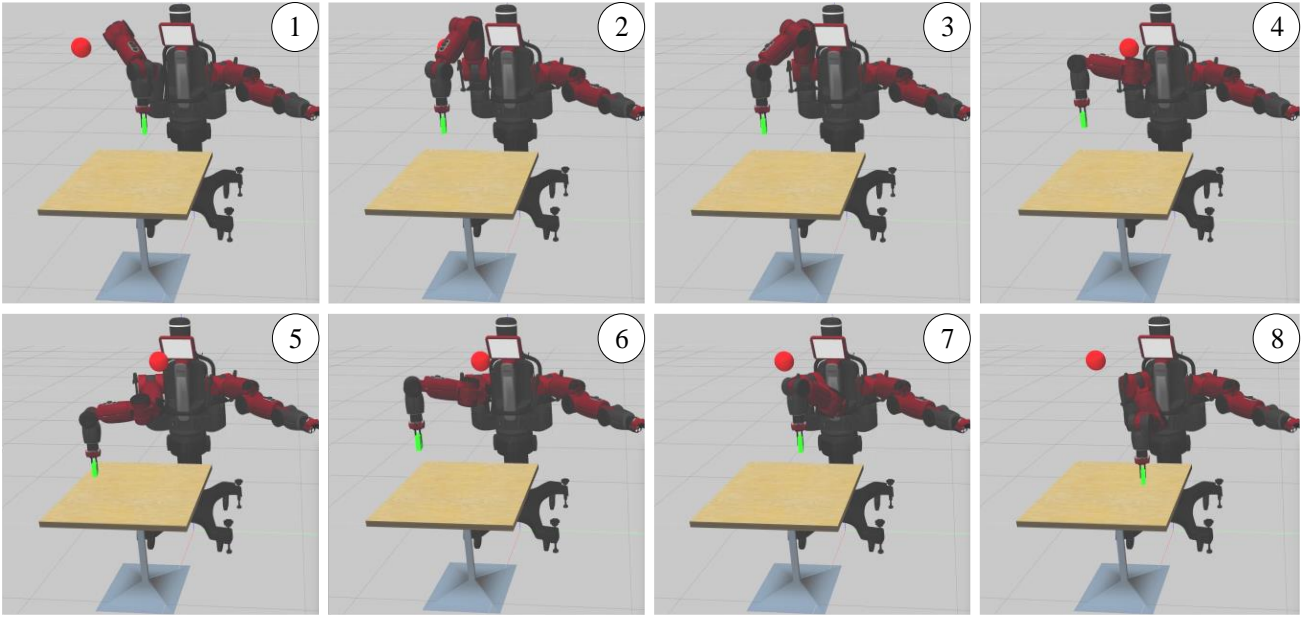

Fig 8. Workflow of the picking & placing use-case. The right arm is in the outward rest position (1) before the movement. There is a dynamic obstacle (red sphere) in the working environment. Our method can generate the inverse kinematics solutions of the right arm (7-DOF) of the Baxter robot.

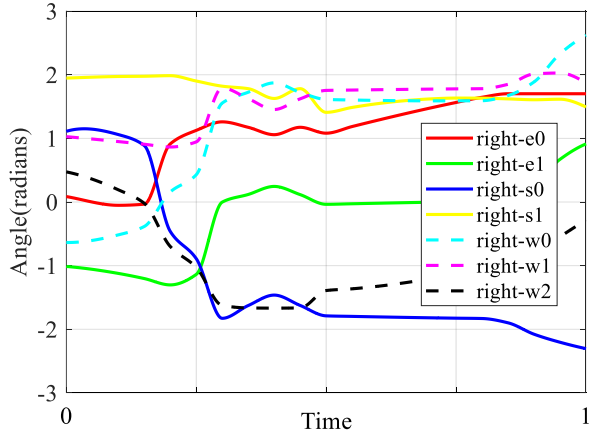

Fig 9. Paths for joints right-e0 to right-w2 via inverse kinematics using the picking & placing use-case in Fig. 8. The arm joints are named in the following manner: e0: Elbow Twist (Roll); e1: Elbow Bend (Pitch); s0: Shoulder Twist (Roll); s1: Shoulder Bend (Pitch); w0: Wrist Twist (Roll); w1: Wrist Bend (Pitch); w2: Wrist Twist (Roll).

$$v_{ms}^{k+1} = v_{ms}^{k} + c_1 r_1^k (p_{ms} - x_{ms}^k) + c_2 r_2^k (g - x_{ms}^k) \quad (15)$$

$$x_{ms}^{k+1} = x_{ms}^{k} + v_{ms}^{k+1} \quad (16)$$

where $x_{ms}^k$ is the current seed value of the joints, the seed value $x_{ms}^{k+1}$ is used to compute the new inverse kinematics solution, $c_1$ and $c_2$ are learning factors that can control the behavior and efficacy of the PSO method and $r_1^k$ and $r_2^k$ are the random numbers (standard uniform distribution number on the open interval (0, 1)). In our formulation, for slight changes to the joints of the redundant manipulator, the termination criterion of PSO is the number of iterations performed or a solution where the adequate value $\psi_{ms}$ is found. The fitness function value $\Phi_{ms}$ can be used to measure the success of the inverse kinematics solution where the RVO constraint $\psi_m$ ($m = 1, \ldots, M$) in the workspace are less than zero.

## VI. IMPLEMENTATION AND PERFORMANCE

In this section, we highlight the performance of our algorithm in simulated environments that contain multiple redundant manipulators in the 3D workspace. We highlight the performance of our method in the dynamic working environment with many moving obstacles, and no assumptions are made about their trajectories. It should be noted that our method is applicable for many manipulators and for multiple dynamic obstacles in the working environment. The general algorithms of the combined TRAC-IK and RVO for the redundant manipulator is shown in Fig. 7.

These simulations are initialized by decomposing the redundant manipulator into a series of spheres. The fitness values are sufficiently small, when the distance between the sphere and the obstacle is sufficiently large. For simplicity, we do not take all the spheres into account while computing the RVO constraints. Therefore, a neighboring region $R_m$ around the current position of sphere $O_m$ is defined and we only consider the spheres of other manipulators and obstacles inside this region (the yellow sphere in Fig. 7). Furthermore, the size of $R_m$ can be determined by the velocity of each sphere and the size of the time-step $\tau$.

We use TRAC-IK to generate the valid configurations and use RVO constraints to define the constraint factor $\psi$ for each configuration. We use the PSO algorithm to relate the inverse kinematics algorithm to the RVO and generate a new inverse kinematics solution based on the $x_{ms}^{k+1}$. The simulation terminates as soon as all the constraints of the fitness function

**Algorithm 1:** IK Solution Generator
(our novel optimization algorithm that accounts for IK and collision avoidance constraints)

**Initialization**: $\psi_{ms}$, $E_{ms}$, $v_{ms}$, $x_{ms}$ ($ms = 1, \ldots, N$)

1 **while** *IK command received* **do**
2   **for** $a = 1$ **to** $T$ (PSO ITERATIONS)
3     **for** $ms = 1$ **to** $N$ (NUMBER OF PSO PARTICLES)
4       $\theta$ (VALID CONFIGURATION) ← TRAC-IK($\theta_{seed} = x_{ms}$)
5       $\psi_{ms}$, $E_{ms}$ ← RVO($\theta$), error($\theta$, $\theta_{LAST\ CONFIGURATION}$)
6       $\Phi_{ms}$ (FITNESS FUNCTION) ← $\alpha\psi_{ms} + \beta E_{ms}$
7       **if** $\psi_m < 0$ (COLLISION) **then**
8         $p_{ms}$, $g$ ← PSO_GET_FROM (($x_1, \ldots, x_N$), $\Phi_{ms}$)
9         $v_{ms}^{k+1}$ ← COM_POSE( $v_{ms}^k$, $p_{ms}$, $x_{ms}^k$, $g$)
10         $x_{ms}^{k+1}$ ← $x_{ms}^k + v_{ms}^{k+1}$
11         $\theta$ (IK SOLUTION) ← TRAC-IK($\theta_{seed} = x_{ms}^{k+1}$)
12 **return** $\theta$

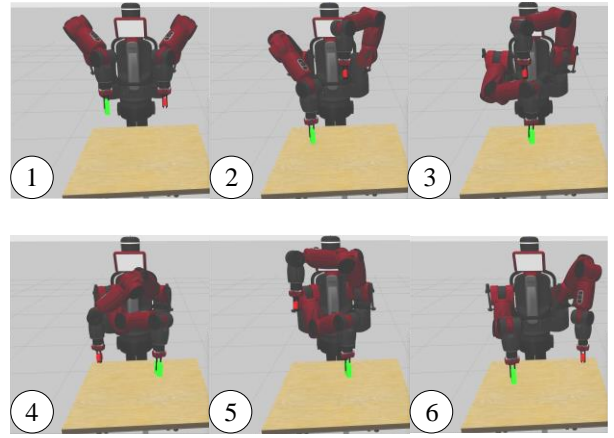

Fig 10. Workflow of the picking & placing use-case of both arms of the Baxter robot. At the beginning of the movement, two arms are in the outward rest position (1). In this simulation (2)-(5), the joint motion command for the right arm was generated first based on our method, and then the motion command for the left arm is triggered and coordinated with the moving right arm. As soon as the right arm moves the green box to the new position, the left arm can complete its place motion (6).

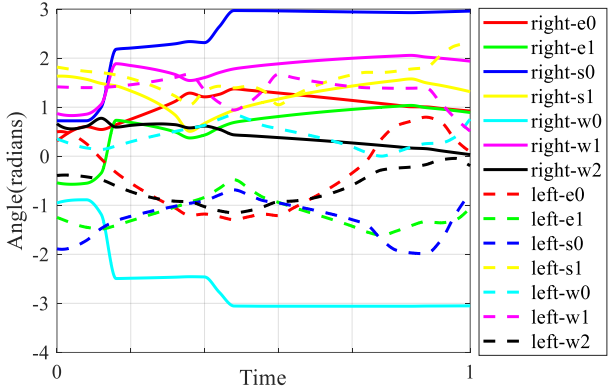

Fig 11. Paths for joints right-e0 to left-w2 (14-DoF) via inverse kinematics using the picking & placing use-case in Fig. 10.

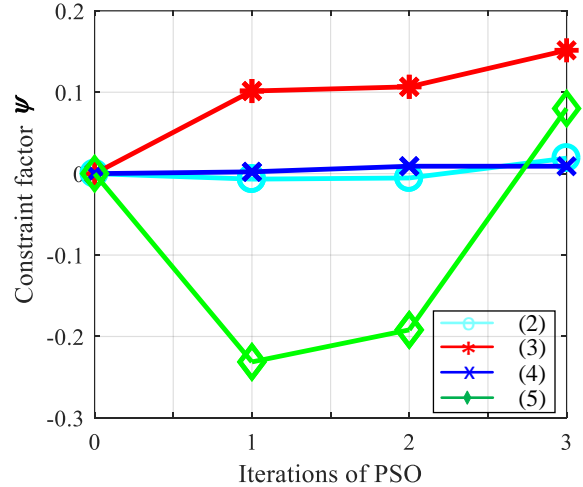

Fig 12. The constraint factor $\psi_{ms}$ of each inverse kinematics solution (Fig.10-(2), (3), (4) and (5)) versus the number of iterations of the PSO.

values $\Phi$ are satisfied by the new inverse kinematics solution. Finally, the inverse kinematics solution is the joint values of the manipulator. The Pseudo-code for the inverse kinematics is presented in Algorithm 1.

Table I shows the results of applying our inverse kinematics method to multiple redundant manipulators in different dynamic environments. The results show that the complexity of the PSO increases as a linear function of the number of manipulators and the dynamic obstacles. Specifically, PSO achieves to maintain the previously best known position of a particle in a swarm, while concurrently searching for new inverse kinematics solutions and improve the results. Our benchmarks show that the overall computation time is directly proportional to the number of DOF, the size of PSO's particles, and the number of iterations of our optimization algorithm.

Our algorithm is implemented in C++ using an ROS Kinetic Kame and Gazebo (Version 7). The code runs on a single notebook CPU i7-7700HQ at 2.81 GHz. The desired end-effector movement directions selected for evaluation are on the right and left arms of the Baxter robot. The end-effector starts from a know position. The configurations of one arms (7-DOF) and the position of the dynamic obstacle are shown in Fig. 8 and 9. The configurations of two arms (14-DOF) are shown in Fig. 10, 11 and 12. Our simulation results demonstrate that our method can deal with multiple redundant manipulators to adjust configurations and avoid collisions. Moreover, all joint angles are within their specified limits.

## VII. CONCLUSION AND FUTURE WORK

We present a novel method for solving the inverse kinematics problem of multiple redundant manipulators by computing collision-free configuration. Our approach combines the inverse kinematics constraints with collision avoidance constraints based on a novel optimization algorithm. To demonstrate the feasibility and efficiency of this method, we have highlighted the performance in our simulator with multiple Baxter robots operating in a shared workspace with moving obstacles. Our algorithm computes a feasible solution in a few milliseconds for multiple robots operating in close proximity. Furthermore, it makes no assumptions about the tasks performed by the manipulators or the dynamic obstacles in the environment.

Our approach has some limitations. The collision avoidance formulation is conservative and it is hard to guarantee that our method can find a feasible solution in all configurations. We would like to evaluate the performance in dynamic scenes with human or other complex obstacles. Furthermore, we can integrate our approach with robots and evaluate their performance in a real-world environment. Another limitation is that the motion planning on the end-effector is not considered in our current formulation.